\documentclass{article}

\PassOptionsToPackage{numbers, compress, sort}{natbib}

 \usepackage[preprint]{neurips_2019}

\usepackage[utf8]{inputenc} 
\usepackage[T1]{fontenc}    
\usepackage{hyperref}       
\usepackage{url}            
\usepackage{booktabs}       
\usepackage{amsfonts}       
\usepackage{nicefrac}       
\usepackage{microtype}      
\usepackage{comment}
\usepackage{microtype}
\usepackage{graphicx}
\usepackage{booktabs} 

\usepackage{times}  
\usepackage{helvet}  
\usepackage{courier}  
\usepackage{url}  
\usepackage{tikz}
\usetikzlibrary{arrows}
\tikzset{>=latex}
\usepackage{amssymb, amsmath}
\usepackage{amsthm}

\newtheorem{definition}{Definition}

\usepackage[font=small]{subcaption}

\usepackage{enumitem}
\usepackage{floatrow}

\newfloatcommand{capbtabbox}{table}[][\FBwidth]

\usepackage{xspace}
\def\reals{\mathbb{R}}
\newcommand{\cpmetric}{\text{\textsc{CPMetric}}\xspace}

\newcommand{\icpd}{\text{\emph{I}-CPD}\xspace}

\newcommand{\ktd}{\text{KTD}\xspace}

\usepackage{color}
\usepackage[textsize=scriptsize]{todonotes}
\definecolor{blue}{RGB}{0, 93, 170}			

\title{\cpmetric: Deep Siamese Networks for Metric Learning on Structured Preferences}

\author{%
 Andrea Loreggia \\
 University of Padova\\
 Padova, Italy\\
 \And
 Nicholas Mattei \\
 Tulane University \\
  New Orleans, LA, USA\\
  \AND
  Francesca Rossi \\
  IBM Research \\
  Yorktown Heights, NY, USA \\
  \And
  K. Brent Venable \\
  Tulane University and IHMC \\
  New Orleans, LA, USA
}

\begin{document}

\maketitle

\begin{abstract}
Preference are central to decision making by both machines and humans.  Representing, learning, and reasoning with preferences is an important area of study both within computer science and across the social sciences.
When working with preferences it is necessary to understand and compute a metric (distance) between sets of objects, e.g., the preferences of two users.
We present \cpmetric, a novel neural network to address the problem of metric learning for structured preference representations.
We use the popular CP-net formalism to represent preferences and then leverage deep neural networks to learn a recently proposed metric function that is computationally hard to compute directly.
\cpmetric is a novel metric learning approach as we learn the Kendal Tau distance between compact representations of partial orders as opposed to the (possibly exponential) induced partial orders.
We find that \cpmetric is able to learn the metric function with high accuracy, outperforming existing approximation algorithms on both the regression and classification tasks using less computation time.
This increased performance over existing direct approximation algorithms persists even when \cpmetric is trained with only a small number of samples compared to the dimension of the solution space, indicating the network generalizes well. 
\end{abstract}

\section{Introduction}

Preferences are central to individual and group decision making by both computer systems and humans.  Due to this central role in decision making the study of representing \cite{RVW11a}, learning \cite{FuHu10a}, and reasoning \cite{DHKP11a,PTV15a} with preferences is a focus of study within computer science and in many other disciplines including psychology and sociology \cite{goldsmith2009preference}. Individuals express their preferences in many different ways: pairwise comparisons, rankings, approvals (likes), positive or negative examples, and many more  examples are collected in various libraries and databases \cite{MaWa13a,MaWa17,BaLi13a}.  A core task in working with preferences is understanding the relationship \emph{between} preferences.  This often takes the form of a dominance query, i.e., which item is more or most preferred, or distance measures, i.e., which object is the closest to my stated preference.  These types of reasoning are important in many domains including recommender systems \cite{pu2011usability,fattah2018cp}, collective decision making \cite{BCELP16a}, and value alignment systems \cite{russell2015research,LoMaRoVe18a,LoMaRoVe18}, among others.

Having a formal structure to model preferences, especially one that directly models dependency, can be useful when reasoning about preferences. For example, it can support reasoning based on inference and causality, and provide more transparency and explainability as the preferences are explicitly represented so the model is scrutable \cite{Kamb19a}. A number of compact preference representation languages have been developed in the literature for representing and reasoning with preferences; see the work of \citet{ABGP16a} for a survey of compact graphical models. In this paper we specifically focus on conditional preference structures (CP-nets) \cite{cpnets}. 

CP-nets are a compact graphical model used to capture qualitative conditional preferences over features (variables) \cite{cpnets}.  Qualitative preferences are an important formalism as there there is experimental evidence that qualitative preferences may more accurately reflect humans' preferences in uncertain information settings \cite{roth1995handbook,ACGM+15a}.  CP-nets are a popular formalism for specifying preferences in the litterature and have been used for a number of applications including recommender systems \cite{pu2011usability} and product specification \cite{fattah2018cp,wang2009web}. Consider a car that is described by values for all its possible features: make, model, color, and stereo options.  A CP-net consists of a dependency graph and a set of statements of the form, \emph{``all else being equal, I prefer x to y.''} For example, in a CP-net one could say \emph{``Given that the car is a Honda Civic, I prefer red to yellow.''}, where condition sets the context for the preference statement over possible alternatives.  These preferences are qualitative as there is no quantity expressing how much I prefer one action over another one.  

A CP-net induces an ordering over all possible  \emph{outcomes}, i.e., all complete assignments to the set of features. This is a partial order if the dependency graph of the CP-net is acyclic, i.e., the conditionality of the statements does not create a cycle, as is often assumed in work with CP-nets \cite{goldsmith2008computational}. The size of the description of the CP-net may be exponentially smaller than the partial order it describes.  Hence, CP-nets are called a \emph{compact} representation and reasoning and learning on the compact structure, instead of the full order, is an important topic of research. Recent work proposes the first formal metric to describe the distance between CP-nets \cite{loreggia2018} and  the related formalism of LP-trees \cite{li2018efficient} in a rigorous way. What is important is not the differences in the surface features of the CP-nets, e.g., a single statement or dependency, but rather the distance between their induced partial orders. Even a small difference in a CP-net could generate a very different partial order, depending on which feature is involved in the modification.  While the metrics proposed by \citet{loreggia2018} are well grounded, they are computationally hard to compute, in general, and approximations must be used.  

Following work in metric learning over structured representations \cite{metricbook,metricarxiv}, we wish to learn the distance between partial orders represented compactly as CP-nets.  We do not want to work with the partial orders directly as they may be exponentially larger than the CP-net representation.  Informally, given two CP-nets, we wish to estimate the distance between their induced partial orders using a neural network.  Notice that this is a fundamentally different task to metric learning over graphs as, although we estimate the distance between graphs (partial orders), we start from a compact representation and not the induced graphs as input.  
There has been recent interest in \emph{deep metric learning} which is similar to the work we consider here.  In deep metric learning we are typically given pairs of input and want to learn an embedding representation of the data that preserves the distance between similar items \cite{sohn2016improved}.  Again, however, this is different from our work as we do not work with individual pairwise comparisons but rather compact structures.  

The aim of this work is not introducing a new graph learning method, an important topic in machine learning \cite{conf/nips/DefferrardBV16,journals/corr/KipfW16}, but rather to merge work in decision theory with machine learning techniques.  This has been done before in the area of \emph{preference learning}, where preferences are inferred from data under a given noise function \cite{FuHu10a}.  However, to our knowledge this is the first attempt to use neural nets to approximate a metric between structured, graphical preference representations. In addition to being an interesting fundamental problem there are practical applications as well. The number of possible CP-nets grows extremely fast, from 481,776 for 4 binary features to over $5.24 \times 10^{40}$ with 7 binary features \cite{AlGoJuMa17}.  However, the computation time of the approximation algorithm proposed by \citet{loreggia2018} scales linearly with the number of features, hence, new methods must be explored.  Therefore, leveraging the inferential properties of neural networks may help us make CP-nets more useful as a preference reasoning formalism.


\smallskip
\noindent
\textbf{Contributions\;} We formalize the problem of metric learning on CP-nets, a compact preference representation, that combines elements of graph embeddings, metric learning, and preference reasoning into one problem. 
We present \cpmetric, a siamese network \cite{siameseNN} trained using pairs of CP-nets represented through their normalized Laplacian matrices and list of cp-statements.  We decompose the problem into two steps: (1) learning a vector representation of the CP-nets and (2) learning the distance metric itself.  We explore the benefits of transfer learning through the use of an autoencoder \cite{hinton2006reducing}.
We evaluate our approach both quantitatively, by judging the accuracy and mean absolute error (MAE) of \cpmetric, and qualitatively, by judging if given two CP-nets we can determine which is closer to a reference point.
\cpmetric is able to learn a good approximation of the distance function and outperforms in terms of both accuracy and speed the current best approximation algorithms on both the regression and classification tasks.  \cpmetric gives good performance even when the network is trained with a small number of samples.

\section{CP-nets}
\label{bkg}

Conditional Preference networks (CP-nets) are a graphical model for compactly representing conditional and qualitative preference relations \cite{cpnets}. CP-nets are comprised of sets of {\em ceteris paribus} preference statements (cp-statements). For instance, the cp-statement, {\em ``I prefer red wine to white wine if meat is served,"} asserts that, given two meals that differ {\em only} in the kind of wine served {\em and} both containing meat, the meal with red wine is preferable to the meal with white wine.  CP-nets have been extensively used in the preference reasoning \cite{dimo,CGMR+13a,RVW11a}, preference learning \cite{chevaleyre2011learning} and social choice \cite{BCELP16a,LangSV,MPRV13a} literature as a formalism for working with qualitative preferences \cite{DHKP11a}.  CP-nets have even been used to compose web services \cite{wang2009web} and other decision aid systems \cite{pu2011usability}.

Formally, a CP-net has a set of features (or variables) $F = \{X_1,\ldots,X_n\}$ with finite domains $\cal D(X_1),\ldots,\cal D(X_n)$. For each feature $X_i$, we are given a set of {\em parent} features $Pa(X_i)$ that can affect the preferences over the values of $X_i$. This defines a {\em dependency graph} in which each node $X_i$ has $Pa(X_i)$ as its immediate predecessors. An {\em acyclic} CP-net is one in which the dependency
graph is acyclic. 
Given this structural information, one needs to specify the preference over the values of each variable $X_i$ for {\em each complete assignment} to the the parent variables, $Pa(X_i)$. This preference is assumed to take the form of a total or partial order over $\cal{D}(X_i)$. A cp-statement for some feature $X_i$ that has parents $Pa(X_i) = \{x_1,\ldots,x_n\}$ and domain $D(X_i) = \{a_1,\ldots,a_m\}$ is a total ordering over $D(X_i)$ and has general form: $x_1=v_1, x_2=v_2, \ldots,x_n=v_{n} : a_1 \succ \ldots \succ a_m$, where for each $X_i \in Pa(X_1): x_i=v_i$ is an assignment to a parent of $X_i$ with $v_i \in \cal{D} (X_i)$. 
The set of cp-statements regarding a certain variable $X_i$ is called the cp-table for $X_i$. 

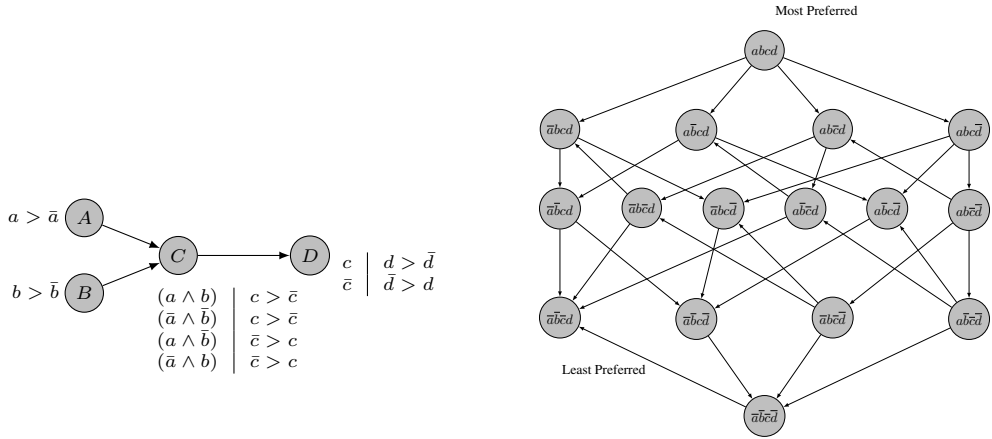
\begin{figure}[h]
\centering
\begin{minipage}[t]{0.48\textwidth}
\scriptsize
\centering
\begin{tikzpicture}[scale=0.5]
\tikzstyle{every node}=[draw,shape=circle,fill=gray!50];
\node (a) at (0, 1) {$A$};
\node (b) at (0, -1) {$B$};
\node (c) at (2.5, 0) {$C$};
\node (d) at (6, 0) {$D$};

\draw [->]  (a) -- (c);
\draw [->]  (b) -- (c);
\draw [->]  (c) -- (d);

\coordinate [label=left: {
\begin{tabular}{c}
$ a > \bar{a} $ \\
\end{tabular}
}] (p) at (0,1);

\coordinate [label=left: {
\begin{tabular}{c}
$ b > \bar{b} $ \\
\end{tabular}
}] (p) at (0,-1);

\coordinate [label=right: {
\begin{tabular}{c|c}
$ (a \wedge b) $					& $ c > \bar{c}$ \\
$ (\bar{a} \wedge \bar{b}) $		& $ c > \bar{c}$ \\
$ (a \wedge \bar{b}) $				& $ \bar{c} > c$ \\
$ (\bar{a} \wedge b) $				& $ \bar{c} > c$ \\
\end{tabular}
}] (p) at (.90,-2.0);

\coordinate [label=right: {
\begin{tabular}{c|c}
$ c $       & $d > \bar{d}$ \\
$ \bar{c} $ &$\bar{d} > d$ \\
\end{tabular}
}] (p) at (6,-0.5);
\end{tikzpicture}
\end{minipage}
\hfill
\begin{minipage}[t]{0.48\textwidth}
\centering
\resizebox{0.9\linewidth}{!}{%
\begin{tikzpicture}[scale=0.3]
\tikzstyle{every node}=[draw,shape=circle,fill=gray!50];
\node[label={[label distance=0.1]20:Most Preferred}] (a) at (0, 0) {$abcd$};
\node[below=of a, xshift=-150] (b) {$\overline{a}bcd$};
\node[below=of a, xshift=-50] (c) {$a\overline{b}cd$};
\node[below=of a, xshift=50] (d) {$ab\overline{c}d$};
\node[below=of a, xshift=150] (e) {$abc\overline{d}$};

\coordinate [below=of b] (bd);
\coordinate [below=of c] (cd);
\coordinate [below=of d] (dd);
\coordinate [below=of e] (ed);

\node[below=of b, xshift=0] (f) {$\overline{a}\overline{b}cd$};
\node[below=of b, xshift=60] (g) {$\overline{a}b\overline{c}d$};
\node[below=of b, xshift=120] (h) {$\overline{a}bc\overline{d}$};
\node[below=of b, xshift=180] (i) {$a\overline{b}\overline{c}d$};
\node[below=of b, xshift=240] (j) {$a\overline{b}c\overline{d}$};
\node[below=of e, xshift=0] (k) {$ab\overline{c}\overline{d}$};

\node[label={[label distance=-0.01]300:Least Preferred},below=of b, yshift=-80] (l) {$\overline{a}\overline{b}\overline{c}d$};
\node[below=of c, yshift=-80] (m) {$\overline{a}\overline{b}c\overline{d}$};
\node[below=of d, yshift=-80] (n) {$\overline{a}b\overline{c}\overline{d}$};
\node[below=of e, yshift=-80] (o) {$a\overline{b}\overline{c}\overline{d}$};

\node[below=of a, yshift=-210] (p) {$\overline{a}\overline{b}\overline{c}\overline{d}$};

\draw [->]  (a) -- (b);
\draw [->]  (a) -- (c);
\draw [->]  (a) -- (d);
\draw [->]  (a) -- (e);

\draw [->]  (b) -- (f);
\draw [->]  (g) -- (b);
\draw [->]  (b) -- (h);

\draw [->]  (c) -- (f);
\draw [->]  (i) -- (c);
\draw [->]  (c) -- (j);

\draw [->]  (d) -- (g);
\draw [->]  (k) -- (d);
\draw [->]  (d) -- (i);

\draw [->]  (e) -- (h);
\draw [->]  (e) -- (k);
\draw [->]  (e) -- (j);

\draw [->]  (f) -- (l);
\draw [->]  (f) -- (m);

\draw [->]  (g) -- (l);
\draw [->]  (n) -- (g);

\draw [->]  (h) -- (m);
\draw [->]  (n) -- (h);

\draw [->]  (i) -- (l);
\draw [->]  (o) -- (i);

\draw [->]  (j) -- (m);
\draw [->]  (o) -- (j);

\draw [->]  (k) -- (n);
\draw [->]  (k) -- (o);

\draw [->]  (p) -- (l);
\draw [->]  (m) -- (p);
\draw [->]  (n) -- (p);
\draw [->]  (o) -- (p);
\end{tikzpicture}
}
\end{minipage}
\caption{A CP-net with $n=4$ features (left) and part of in the induced partial order (right).  Note that the partial order is over all $2^n = 16$ possible combinations and arrows denote the dominance relation.  We have arranged the nodes so that each is one flip between the levels.}
\label{excpnet1}
\end{figure}

Consider the CP-net depicted graphically in Figure \ref{excpnet1} (left) with features are $A$, $B$, $C$, and $D$.  Each variable has binary domain containing $f$ and $\overline{f}$ if $F$ is the name of the feature.  All cp-statements in the CP-net are: $a \succ \overline{a}$, $b \succ \overline{b}$, $(a \wedge b) : c \succ \overline{c}$, $(\overline{a} \wedge \overline{b}) : c \succ \overline{c}$, $(a \wedge \overline{b}) : \overline{c} \succ c$, $(\overline{a} \wedge b) : \overline{c} \succ c$, $c: d \succ \overline{d}$, $\overline{c}: \overline{d} \succ d$. Here, statement $a \succ \overline{a}$ represents the unconditional preference for $A=a$ over $A=\overline{a}$, while statement $c: d \succ \overline{d}$ states that $D=d$ is preferred to $D=\overline{d}$, given that $C=c$.  
The semantics of CP-nets depends on the notion of a {\em worsening flip}: a change in the value of a variable to a less preferred value according to the cp-statement for that variable. For example, in the CP-net above, passing from $abcd$ to $\overline{a}bcd$ is a worsening flip since $c$ is better than $\overline{c}$ given $a$ and $b$. One outcome $\alpha$ is {\em preferred to} or \emph{dominates} another outcome $\beta$ (written $\alpha \succ \beta$) if and only if there is a chain of worsening flips from $\alpha$ to $\beta$. This definition induces a preorder over the outcomes, which is a partial order if the CP-net is acyclic \cite{cpnets}, as depicted in Figure \ref{excpnet1} (right).  

The complexity of dominance and consistency testing in CP-nets is an area of active study in preference reasoning \cite{goldsmith2008computational,RVW11a}. Finding the optimal outcome of a CP-net is NP-hard~\cite{cpnets} in general but can be found in polynomial time for acyclic CP-nets by assigning the most preferred value for each cp-table.  Indeed, acyclic CP-nets induce a lattice over the outcomes as (partially) depicted in Figure \ref{excpnet1} (right).  The induced preference ordering, Figure \ref{excpnet1} (right), can be exponentially larger than the CP-net Figure \ref{excpnet1} (left), which motivates learning a metric using only the (more compact) CP-net.



\section{Metric Learning on CP-nets}

Metric learning algorithms aim to learn a metric (or distance function) over a set of training points or samples \cite{sohn2016improved}. The importance of metrics has grown in recent years with the use of these functions in many different domains: from clustering to information retrieval and from recommender systems to preference aggregation. For instance, many clustering algorithms like the $k$-Means or classification algorithm including $k$-Nearest Neighbor use a distance value between points \cite{nnpattern,lsquant}. In many recommender systems a similarity function allows for a better profiling \cite{web-services}.

Formally, a metric space is a pair $(M,d)$ where $M$ is a set of elements and $d$ is a function $d : M \times M \rightarrow \reals$ where $d$ satisfies four criteria.  Given any three elements $A,B,C \in M$, $d$ must satisfy: (1) $d(A,B) \geq 0$, there must be a value for all pairs; (2) $d(A,B) = d(B,A)$, $d$ must be symmetric; (3) $d(A,B) \leq d(A,C) + d(C,B)$; $d$ must satisfy the triangle inequality; and (4) $d(A,B) = 0$ if and only if $A = B$; $d$ can be zero if and only if the two elements are the same.

\citet{xing_2002} first formalized the problem of metric learning, i.e., learning the metric directly from samples rather than formally specifying the function $d$. This approach requires training data, meaning that we have some oracle that is able to give the value of the metric for each pair. The success of deep learning in many different domains \cite{chopra2005learning,krizhevsky2012imagenet} has lead many researchers to apply these approaches to the field of metric learning, resulting in a number of important results \cite{metricbook,metricarxiv,sohn2016improved}.

In this work we focus on metric spaces ($M$, $d$) where $M$ is a set of CP-nets.  Given this, we want to learn the distance $d$ which best approximates the Kendall tau distance (\ktd) \cite{Kendall} between the induced partial orders.  Informally, the Kendall tau distance between two orderings is the number of pairs that are \emph{discordant}, i.e., not ordered in the same way, in both orderings.  This distance metric extended to partial orders (Definition \ref{ktd}) was defined and proved to be a metric on the space of CP-nets by \citet{loreggia2018}.  To extend the classic \ktd to CP-nets a penalty parameter $p$ defined for partial rankings \cite{fagin2006} was extended to the case of partial orders. \citet{loreggia2018} assume that all CP-nets are acyclic and in minimal (non-degenerate) form, i.e., all arcs in the dependency graph have a real dependency expressed in the cp-statements, a standard assumption in the CP-net literature (see e.g., \cite{AGJM+15a,AlGoJuMa17,cpnets}).

\begin{definition}
\label{ktd}
Given two CP-nets $A$ and $B$ inducing partial orders 
$P$ and $Q$ over the same unordered set of outcomes $U$:
$
KTD(A,B) = KT(P,Q)=\sum_{\forall i,j \in U, i \neq j} K^{p}_{i,j}(P,Q)
$
\noindent
where $i$ and $j$ are two outcomes with $i \neq j$ (i.e., iterate over all unique pairs), we have:
\begin{enumerate}[itemsep=-0.25em]
\item  $K^{p}_{i,j}(P,Q)=0$ if $i,j$ are ordered in the same way
or are incomparable in $P$ and $Q$;
\item $K^{p}_{i,j}(P,Q)=1$ if $i,j$ are
ordered inversely in $P$ and $Q$;
\item $K^{p}_{i,j}(P,Q)=p$, $0.5 \leq p <1$  if $i,j$ are ordered in
$P$ and incomparable in $Q$ (resp. $Q,P$).
\end{enumerate}

To make this distance scale invariant, i.e., a value in $[0,1]$, it is divided by $|U|$.\end{definition}

CP-nets present two important and interesting challenges when used for metric learning.  The first is that we are attempting to learn a metric via a compact representation of a partial order.  We are not learning over the partial orders induced by the CP-nets directly, as they could be exponentially larger than the CP-nets.  The second challenge is the encoding of the graphical structure itself.  Graph learning with neural networks is still a active and open area of research \cite{journals/corr/BrunaZSL13,henaff2015convolutional,conf/nips/DefferrardBV16} including the popular Graph Convolutional Neural Network (GraphGCN) \cite{journals/corr/KipfW16} and methods to speed up graph learning \cite{CMX18a}. \citet{DBLP:journals/corr/GoyalF17} give a complete survey of recent work as well as a Python library of implementations for many of these techniques.  Most of these works focus on finding good embeddings for the nodes of the network and then using collections of these learned embeddings to represent the graph for, e.g., particular segmentation or link prediction tasks. None of these techniques have been applied to embedding graphs for metric learning.

\section{Structure of \cpmetric}

\begin{figure}[h!]
\begin{center}
\includegraphics[width=0.70\linewidth]{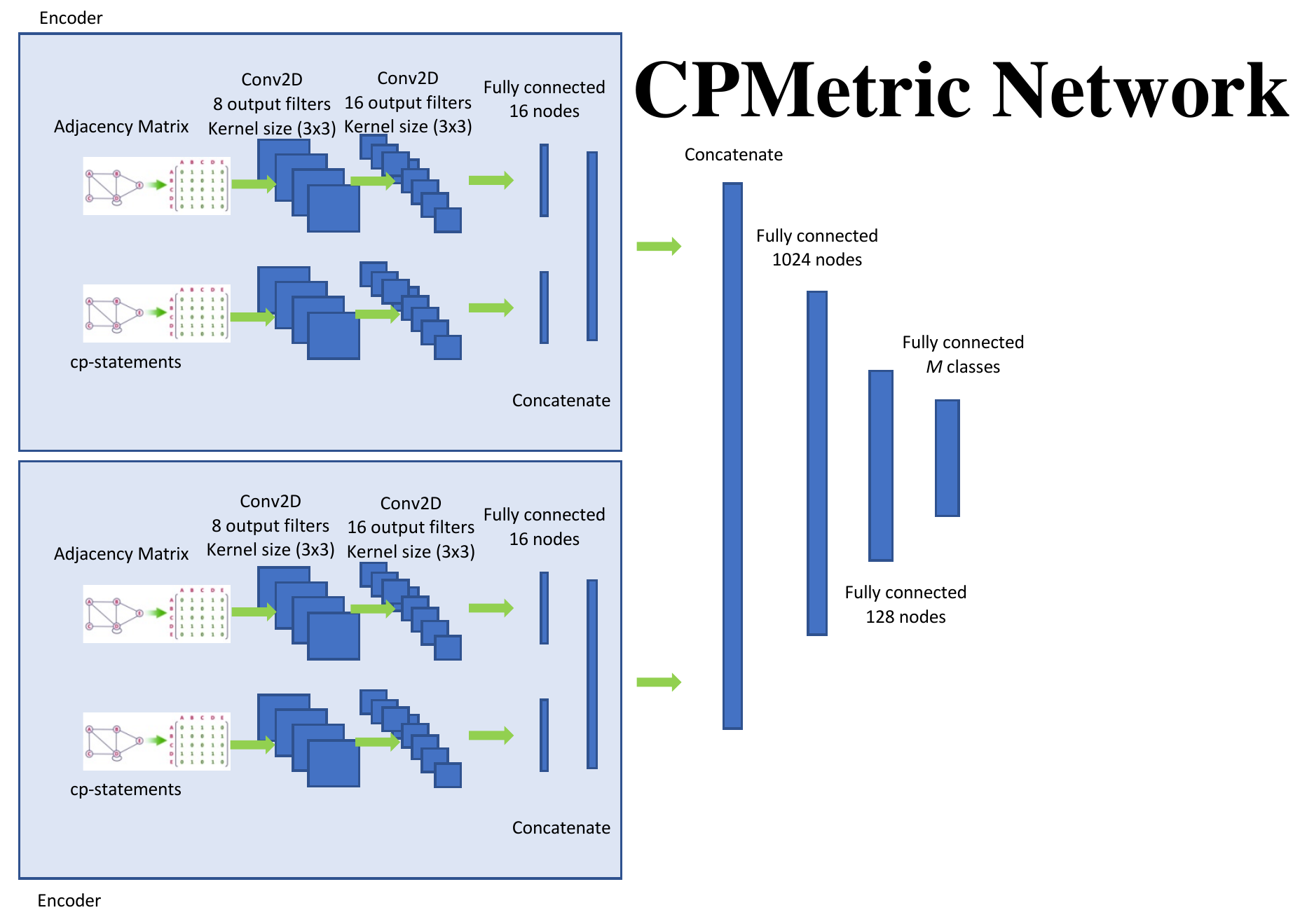}
\end{center}
\caption{Structure of \cpmetric: CP-nets are provided to the encoder as a normalized Laplacian matrix and a list of cp-statements. The encoders output a compact representation of the CP-nets which is then concatenated and passed to the fully connected layers that connect to an $m$ class classifier over  $[0,1]$ to predict \ktd.  For the regression task the network structure is the same except we change the output layer to be one node with a softmax activation layer.}
\label{fig:cpdist}
\end{figure}

\begin{figure}[h]
\begin{center}
\includegraphics[width=0.90\linewidth]{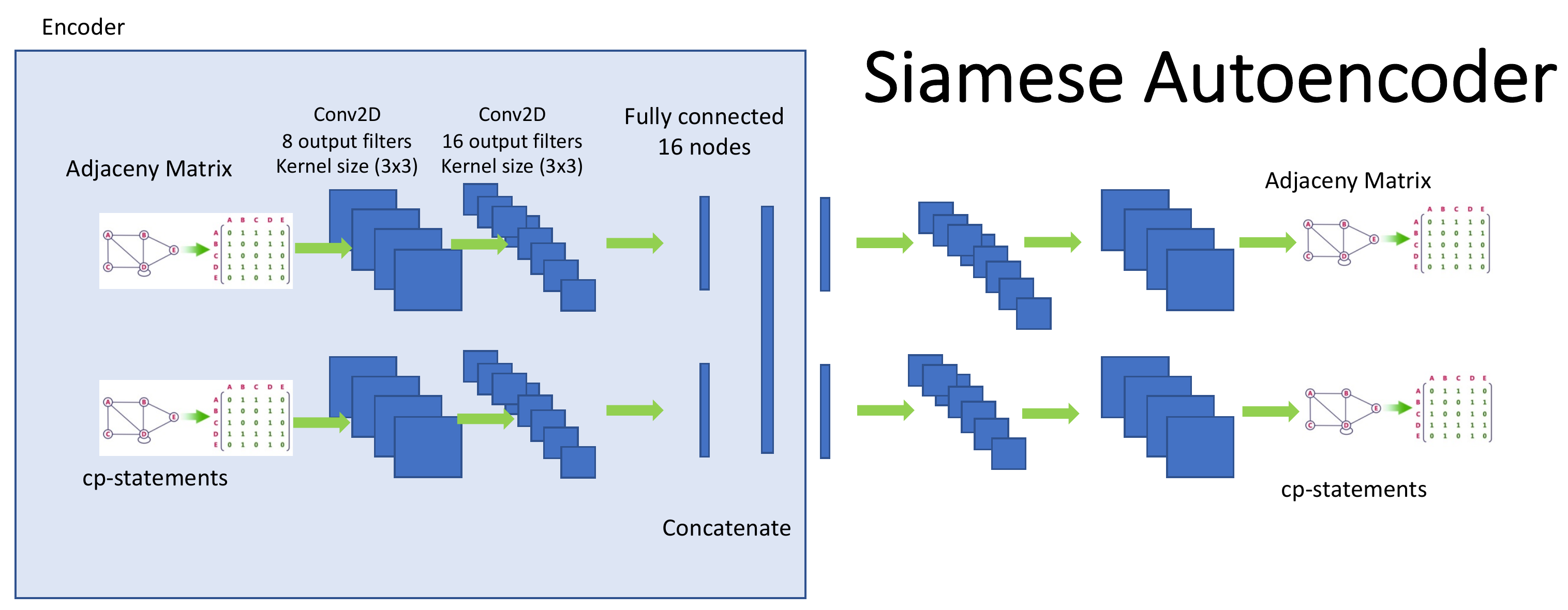}
\end{center}
\caption{Structure of \emph{Siamese Autoencoder}: this version of the autoencoder uses a combined representation for the adjency matrix and the cp-statements.}
\label{fig:autoencoder}
\end{figure}

The architecture of \cpmetric is depicted in Figure \ref{fig:cpdist}. In this section we will discuss the encoding used for the CP-nets and the design of our autoencoders, depicted in Figure \ref{fig:autoencoder} that are used for transfer learning in this domain.  We would like to leverage transfer learning in this domain since training examples become prohibitively expensive to compute at higher values of $n$ as computing $\ktd$ requires exponential time in the size of the CP-net.  Hence, if we can learn a good encoding for CP-nets it may be possible to train a network for small $n$ and use it for problems with larger CP-nets.

In our task the metric space is $(M,d)$ where $M$ is a set of compact, graphical preferences that induce a partial order and our goal is to learn the metric $d$ \emph{only} from the compact, graphical representation.  The key challenge is the need to find a vector representation of not only the graph but the cp-statement. We represent a CP-net $I$ over $m$ using two matrices.  First is the adjacency matrix $adj_I$ which represents the dependency graph of the CP-net and is a $m \times m$ matrix of 0s and 1s. 
The second matrix represents the list of cp-statements $cpt_I$, which is a $m \times 2^{m-1}$ matrix, where each row represents a variable $X_i \in F$  and each column represents a complete assignment for each of the variables in $F \setminus X_i$. The list is built following a topological ordering of variables in the CP-net. Each cell $cpt_{I}(i,j)$ stores the preference value for the $i$th variable given the $j$th assignment to variables in $F \setminus X_i$.

In graph learning, the central research question is how to redefine operators, such as convolution and pooling, so as to generalize convolutional neural network (CNN) to graphs \cite{henaff2015convolutional,conf/nips/DefferrardBV16}. The most promising research uses a spectral formulation of the problem \cite{journals/spm/ShumanNFOV13,journals/corr/BrunaZSL13}.  The issue is that networks are sensitive to isomorphisms of the adjacency matrix, hence directly using an adjacency matrix would result in a siamese network that would not recognize isomorphic structures. We follow in the spirit of the work by \citet{journals/corr/KipfW16} for GCN and use a simple convolutional network structure removing pooling layers from \cpmetric, as we do not define any pooling operator over the graph structure. In graph spectral analysis, the Laplacian matrix is preferred as it has better properties for encoding, e.g., density, compared to just the adjacency matrix. The Laplacian matrix $L = D-A$, where $D$ is the degree matrix, a diagonal matrix whose $i$th diagonal element $d_i$ is equal to the sum of the weights of all the edges incident to vertex $i$, and $A$ is the adjacency matrix representing the graph. The normalized Laplacian $\mathcal{L} = I - D^{\frac{1}{2}} \times A \times D^{\frac{1}{2}}$ \cite{journals/spm/ShumanNFOV13}.  While the Laplacian matrix is still susceptible to exchanges of rows or columns, its spectrum (the vector of its eigenvalues) is an isomorphism invariant of a graph. The same graph can be represented using different structures (and this can be seen as a data augmentation technique) and we need all of these structures to learn the metric, so we cannot collapse to a single spectrum representation of the graph.

The set of training examples $X=\{x_1,\ldots, x_n\}$ is made up of pairs of CP-nets represented through their normalized Laplacians and the cp-statements. The set of corresponding labels $Y=\{y_1,\ldots, y_n\}^T$, where each $y_i \in Y, y_i \in [0,1]$  is the normalized value of \ktd between the CP-nets in $x_i$. Each $x_i \in X$ is then a tuple $(\mathcal{L}_A, cpt_A, \mathcal{L}_B, cpt_B)$ representing a pair of CP-net $(A,B)$ by their Laplacian, $\mathcal{L}_A$, and the encoding of their cp-statements, $cpt_A$.

The purpose of the two input components of \cpmetric, labeled \emph{Encoder} in Figure \ref{fig:cpdist}, is to output a compact representation of CP-nets. To improve performance with networks of this structure, a well-established practice is to train an autoencoder separately, and then transfer the weights to the main network \cite{hinton2006reducing,lecun95convolutional}. We will evaluate two different approaches to transfer learning in our setting. First, we use two different autoencoders: one for the normalized Laplacian matrix and the other for the cp-statements. The two autoencoders are trained separately and then weights are transferred to the main network.  We denote this approach as \emph{Autoencoder} in subsequent experiments. In the second approach, shown in Figure \ref{fig:autoencoder} and denoted as \emph{Siam. Autoencoder}, we use a unique autoencoder designed to combine the two components of CP-nets.  Informally, the output of two encoders are concatenated and then split into their respective components to be decoded. We conjecture that this combination should allow more information about the CP-net to be used.

\section{Experiments}
We train \cpmetric to learn the \ktd metric, varying the number of features of the CP-nets $n \in \{3, \ldots, 7\}$ and using two different autoencoder designs. We evaluate our networks on both the regression and classification tasks and measure their performance against the current best approximation algorithm, \icpd \cite{loreggia2018}, for computing the \ktd between two CP-nets.    In the regression task the network computes the distance value exactly while in the classification task we divide the output in $m=10$ intervals and the network must select the correct interval. 

\subsection{Data Generation and Training}

\begin{figure*}
\begin{floatrow}
\ffigbox{%
\centering
\includegraphics[width=.9\linewidth]{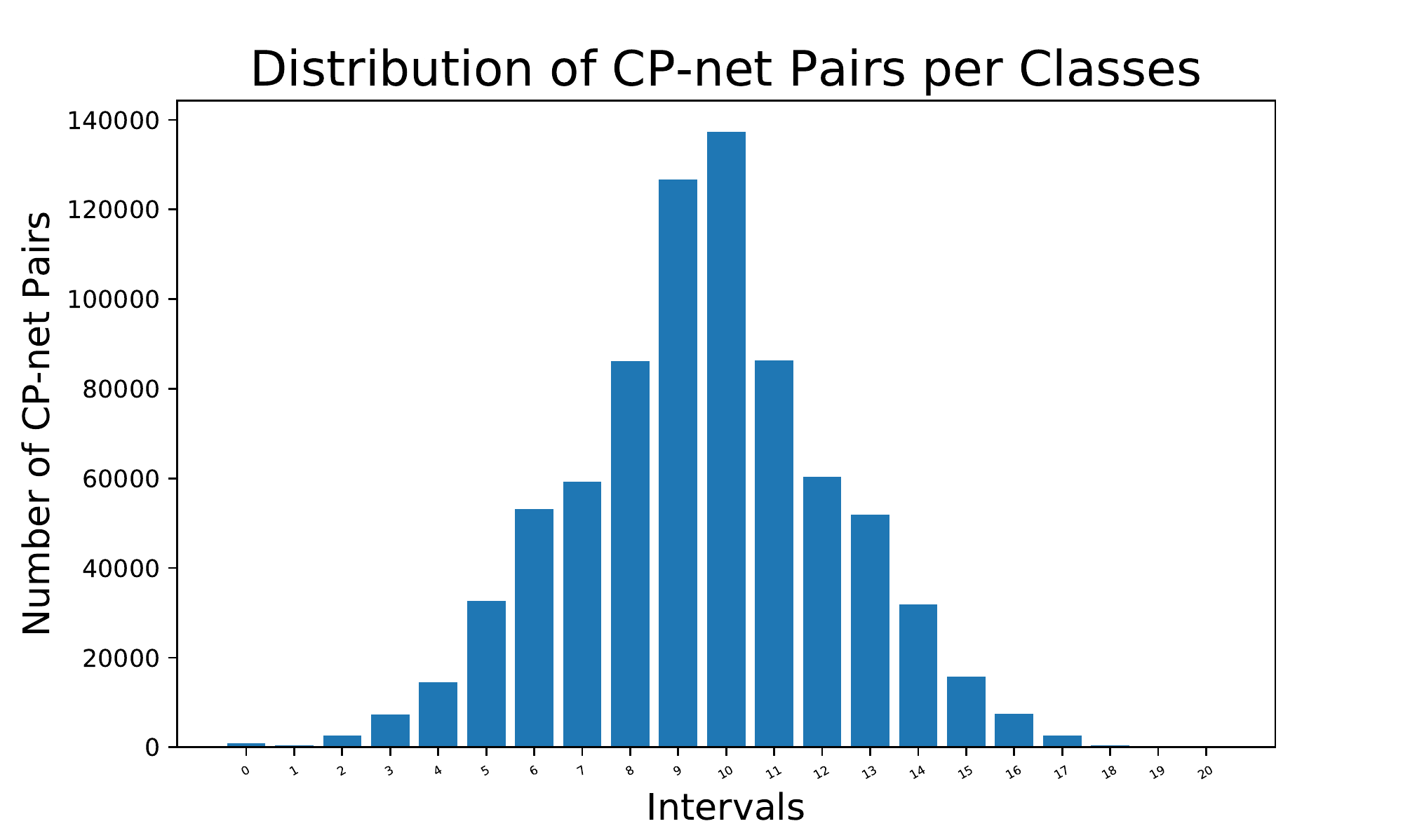}
}{%
\caption{Histogram of the number of CP-net pairs per interval across all experimental datasets. CP-nets pairs are not distributed uniformly in the class intervals.}
\label{fig:dist}
}
\capbtabbox{%
\centering
\resizebox{0.95\linewidth}{!}{
\begin{tabular}{lcc}  
\toprule
N    & \icpd & Autoencoder Neural Network\\
\midrule
3	&  0.69 (0.48) msec & 0.087 (0.004) msec \\
4	&  1.09 (0.33) msec & 0.098 (0.004) msec \\
5	&  1.85 (0.49) msec & 0.100 (0.005) msec \\
6	&  3.16 (0.74) msec & 0.114 (0.003) msec \\
7	&  4.65 (0.86) msec & 0.138 (0.001) msec \\
\bottomrule
\end{tabular}
}
}{%
\caption{Comparison of the mean runtime for a single triple over 1000 trials on the qualitative comparison task of the neural network and \icpd \cite{loreggia2018}.}
\label{tab:time}
}
\end{floatrow}
\end{figure*}

For each number of features $n \in \{3, \ldots, 7\}$ we generate 1000 CP-nets uniformly at random using the generators from Allen et al. \cite{AGJM+15a,AlGoJuMa17}. This set of CP-nets is split into a training-generative-set (900 CP-nets) and test-generative-set (100 CP-nets) 10 different ways to give us 10 fold cross validation. For each fold we compute the training and test dataset comprised of all, e.g., $900 \choose 2$, possible pairs of CP-nets from the training-generative-set and test-generative-set, respectively, along with the value of \ktd for that pair. While we generate the CP-nets themselves uniformly at random observe that this creates an unbalanced set of distances -- it induces a normal distribution -- and hence our sets are unbalanced. Figure \ref{fig:dist} shows the distribution of of CP-net pairs over 20 intervals for all CP-nets generated for $n \in \{3, \ldots, 7\}$.  While our classification experiments are for $m=10$ classes, dividing the interval into 20 classes provides a better visualization of the challenge of obtaining training samples at the edges of the distribution. 

We ran a preliminary experiment on balancing our dataset by sub-sampling the training and test datasets. In these small experiments, performance was much worse than performance on the unbalanced dataset, e.g., for classification the MAE for $n=3$ was $0.626$ and $n=4$ was $0.4962$ versus $0.2734$ and $0.2548$ for the unbalanced results (Figure \ref{tab:classification}). Because we are learning a metric, for each CP-net $A$, there is only one CP-net $B$ such $KTD(A,B)=1$ and only one CP-net $C$ such $KTD(A,C)=0$.  Consequently, attempting to balance or hold out CP-nets from test or train can lead to poor performance.  We conjecture that in order to improve this task we should perform some kind of data augmentation, but this would introduce more subjective assumptions on how and where data should be augmented \cite{wong2016understanding}.

All training was done on a machine with 2 x Intel(R) Xeon(R) CPU E5-2670 @ 2.60GHz and one NVidia K20 128GB GPU.  We train \cpmetric for 70 epochs using the Adam optimizer \cite{KinBa17}. For each number of features of the CP-net $n$ we use all ${900 \choose 2}$ pairs in the training-set.  There are only 488 binary CP-nets with 3 features \cite{AlGoJuMa17}, hence, for $n=3$ the training-set is 17K samples while for $n>3$ the number of samples in the training-set is 800K. Both the \emph{Autoencoder} and \emph{Siamese Autoencoder} models are trained for 100 epochs using the Adam optimizer \cite{KinBa17} using the same training-set. Model weights from the best performing epoch are saved and subsequently transferred to the deep neural network used to learn the distance function.

\begin{figure*}[h!]
\centering
\begin{subfigure}{0.45\linewidth}
	\centering
	\includegraphics[width=\linewidth]{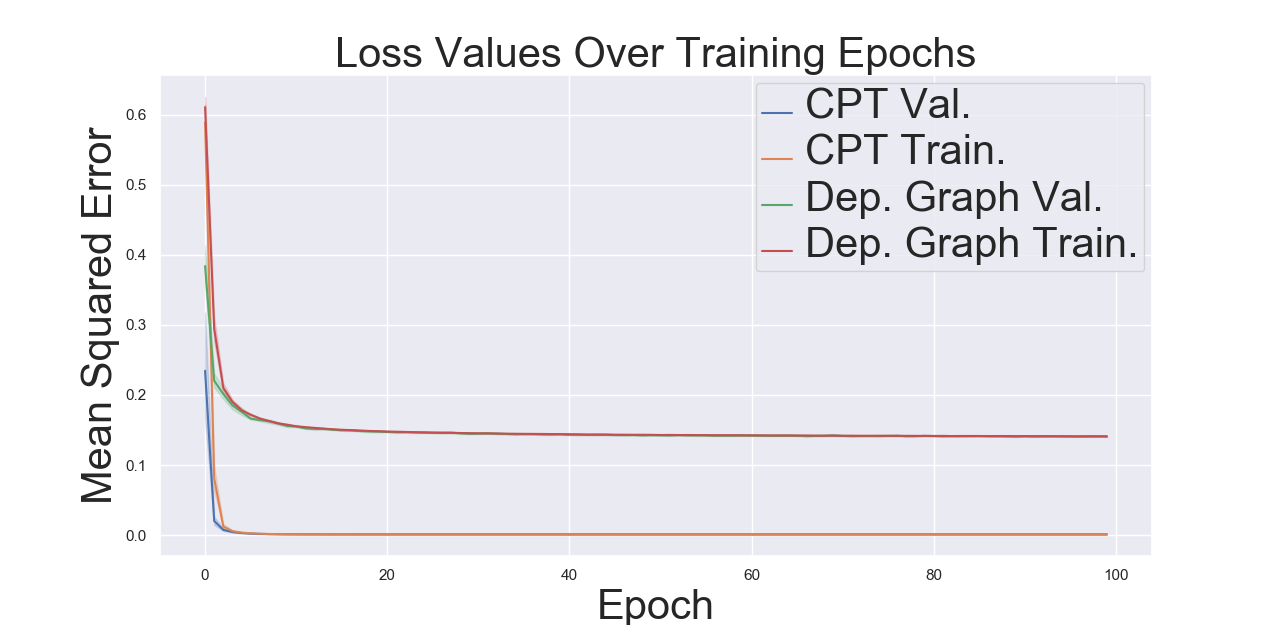}
	\caption{Autoencoder loss for 100 epochs.}
	\label{fig:train}
\end{subfigure}
\hfill
\begin{subfigure}{0.45\linewidth} 
	\centering
	\includegraphics[width=\linewidth]{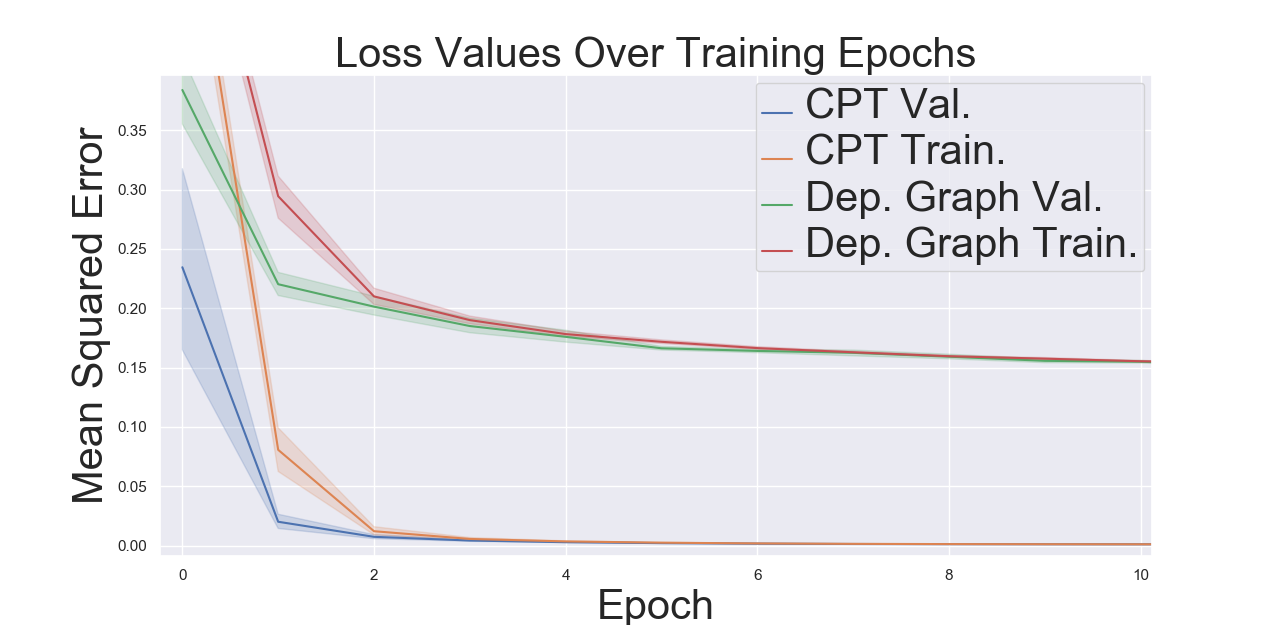}
	\caption{Autoencoder loss for 10 epochs.}
	\label{fig:zoomed}
\end{subfigure}
\caption{Performance of the autoencoder on the validation and training set across epochs.  Note }
\label{fig:autoencoder_graph}
\end{figure*}

The training and validation loss for the autoencoder is shown in Figure \ref{fig:autoencoder_graph}.  Observe that the loss for the CPT representation approaches zero after only 3 epochs for both the training and validation phases.  The same trend is true for the adjacency matrix, though the loss converges to $\approx 0.15$.

\subsection{Quantitative Performance: Classification and Regression}

The first task for \cpmetric is classifying the distance between two CP-nets, $A$ and $B$, into the same one of $m=10$ intervals of $[0,1]$ where the value of \ktd lies.  Table \ref{tab:classification} gives the F-score, Cohen's Kappa (Cohen-$\kappa$) \cite{Cohe95a}, and mean absolute error (MAE) for the task with no autoencoder and each of the two autoencoder variants.  Cohen's $\kappa$ is a measure of inter-rater agreement where the two raters are the particular instance of \cpmetric and the actual value of \ktd.  We measure mean absolute error as a value over the number of intervals between the value returned by \cpmetric and \ktd.  For example, a MAE of 1.0 means that \cpmetric is off by one interval, on average. In this setting, using a random classifier to guess the interval with $m=10$ possible intervals and a normal distribution like the one seen in Figure \ref{fig:dist} would give an F-score $\approx 0.19$.

\begin{table*}[h!]
\centering
\caption{ \small Performance of \cpmetric on the classification task with and without the autoencoders.  Numbers in parenthesis are standard deviations.  Mean absolute error is computed as the number of intervals between the true and predicted values for the classification task.}
\resizebox{\linewidth}{!}{
\begin{tabular}{lcccccccccc}  
\toprule
& \multicolumn{3}{c}{No Autoencoder} & \multicolumn{3}{c}{Autoencoder} & \multicolumn{3}{c}{Siam. Autoencoder} & \icpd \\ \cmidrule(r){2-11} 
N    	& F-score 		& Cohen-$\kappa$ 	& MAE 		& F-score 		& Cohen-$\kappa$ 	& MAE 		& F-score 		& Cohen-$\kappa$ 	& MAE & MAE \\ 
\midrule
3	& 0.6643 (0.0275)	& 0.6113	& 0.3449	& 0.7051 (0.0306)	& 0.6578	& 0.2986	& 0.7295 (0.0501)	& 0.6860	& \textbf{0.2734} & 0.4235 \\
4	& 0.7424 (0.0096)	& 0.6762	& 0.2582	& 0.7483 (0.0085)	& 0.6824	& \textbf{0.2525}	& 0.7459 (0.0088)	& 0.6796	& 0.2548  & 0.4515 \\
5	& 0.7074 (0.0111)	& 0.6146	& 0.3015	& 0.7271 (0.0084)	& 0.6385	& 0.2833	& 0.7278 (0.0077)	& 0.6393	& \textbf{0.2831} & 0.3875\\
6	& 0.6945 (0.0130)	& 0.5799	& 0.3194	& 0.7157 (0.0198)	& 0.6073	& 0.2971	& 0.7161 (0.0141)	& 0.6081	& \textbf{0.2969} & 0.3645 \\
7	& 0.6887 (0.0227)	& 0.5571	& 0.3256	& 0.6497 (0.0892)	& 0.4957	& 0.3830	& 0.6884 (0.0274)	& 0.5549	& \textbf{0.3266} & 0.3340\\
\bottomrule
\end{tabular}
}
\label{tab:classification}
\end{table*}

\begin{table*}[h!]
\centering
\caption{\small MAE of \cpmetric on the regression task with and without the autoencoders.  MAE is the mean over 10 folds and numbers in parenthesis are the standard deviations.}
\resizebox{0.55\linewidth}{!}{
\begin{tabular}{lcccc}  
\toprule
& No Autoencoder & Autoencoder & Siam. Autoencoder & \icpd \\ 
\midrule
3	& 0.0470 & 0.0426 &\textbf{0.0421} & 0.0576 \\
4	& 0.0248 (0.0008) & \textbf{0.0242 (0.0005)} & 0.0243 (0.0007) & 0.0526 \\
5	& 0.0269 (0.0006) & \textbf{0.0261 (0.0008)} & 0.0262 (0.0008) & 0.0463 \\
6	& 0.0257 (0.0007) & \textbf{0.0255 (0.0007)} & 0.0256 (0.0006) & 0.0405 \\
7	& 0.0257 (0.0008) & 0.0257 (0.0022) & \textbf{0.0252 (0.0015)} & 0.0373 \\
\bottomrule
\end{tabular}
}
\label{tab:regression}
\end{table*}

Looking at Table \ref{tab:classification} we see that \cpmetric achieves outperforms  the \icpd approximation algorithm across the test instances. The overall accuracy, measured as F-score, is above 70\% across all CP-net sizes and we see that on average it is off by less than 0.5 intervals as measured by the MAE.  
The values for Cohen's $\kappa$ indicate good agreement between the two methods and this is borne out by high accuracy numbers.  
The most interesting overall effect in Table \ref{tab:classification} is that the performance does not decay much as we increase the number of features.  Indeed, the F-score remains very stable across the range. We interpret this to mean that \cpmetric is learning a good generalization of the distance function even when the solution space is exponentially larger than the number of training examples.

%

Table \ref{tab:regression} we see the results of the much harder regression task. Again we see that \cpmetric is able to out perform the state of the art \icpd approximation across the board.  While for $n=3$ the values are similar, for $n \in \{4, \ldots 7\}$ \cpmetric is giving a $\approx 30\%$ decrease in error, $\approx 0.015$ absolute decrease.  Looking at results from Table \ref{tab:time} we can see that \cpmetric is doing this significantly faster than \icpd as well.  It is interesting to note that in Table \ref{tab:regression} all versions of our network are outperforming \icpd, whether or not we first train the autoencoder. 

Turning to the question of transfer learning for this task we see that the use of the autoencoders strictly increases the performance of the network on the classification and regression task.  In both cases the best performing networks use one of the two autoencoder variants we tested.  The \emph{Siamese Autoencoder} slightly out performs the plain \emph{Autoencoder} when looking at MAE for the classification task, though the results are more mixed for F-score and Cohen-$\kappa$.  In the regression task the \emph{Siamese Autoencoder} is better at the end points and the two networks are statistically indistinguishable for $n \in \{4,5,6\}$.  These results indicate that the use of an autoencoder can significantly help in this task, though the exact design of that autoencoder remains an important question for future work.  Important future work is using an autoencoder trained for a smaller number of features to bootstrap learning for larger numbers of features.

\section{Conclusion}
We present \cpmetric, a novel neural network model to learn a metric (distance) function between partial orders induced from a CP-net, a compact, structured preference representation.  To our knowledge this is the first use of neural networks to learn structured preference representations.  We leverage recent research in metric learning and graph embeddings to achieve state of the art results on the task.  We also demonstrate the value of transfer learning in this domain through the use of two novel autoencoders for the CP-net formalism.  Important directions for future work include integrating novel graph learning techniques to our networks and extending our work to other formalisms including, e.g.,  PCP-nets \cite{CGMR+13a} and LP-trees \cite{li2018efficient}.  PCP-nets are a particularly interesting direction as they have been proposed as an efficient way to model uncertainty over the preferences of a single or multiple agents \cite{CGGM+15a}

%

{\small
\bibliographystyle{plainnat}

}

\end{document}